\documentclass{article}

\usepackage{arxiv}

\usepackage[utf8]{inputenc} 
\usepackage[T1]{fontenc}    
\usepackage{hyperref}       
\usepackage{url}            
\usepackage{booktabs}       
\usepackage{amsfonts}       
\usepackage{nicefrac}       
\usepackage{microtype}      
\usepackage{lipsum}
\usepackage{graphicx}

\usepackage{amsmath}
\usepackage{xcolor}
\usepackage{algorithm}
\usepackage{algorithmic}
\usepackage{float}  

\usepackage[backend=biber,
  style=numeric,
  sorting=none,
  maxnames=2,
  minnames=2,
  giveninits=true
]{biblatex}

\DeclareFieldFormat{titlecase}{\MakeSentenceCase*{#1}}

\title{Peregrine:  One-Shot Fine-Tuning for FHE Inference of General Deep CNNs}

\author{
  Huaming Ling \\
  Cipherflow \\
  Shenzhen, China \\
  \And
  Ying Wang \\
  Cipherflow \\
  Shenzhen, China \\
  \And
  Si Chen \\
  Cipherflow \\
  Shenzhen, China \\
  \And 
  Junfeng Fan \\
  Open Security Research \\
  Shenzhen, China
  \AND
  \texttt{\{huaming.ling,ying.wang,si.chen,fan\}@osr-tech.com}
}

\begin{document}
\maketitle

\begin{abstract}
We address two fundamental challenges in adapting general deep CNNs for FHE-based inference: approximating non-linear activations such as ReLU with low-degree polynomials while minimizing accuracy degradation, and overcoming the ciphertext capacity barrier that constrains high-resolution image processing on FHE inference.
Our contributions are twofold:
(1) a single-stage fine-tuning (SFT) strategy that directly converts pre-trained CNNs into FHE-friendly forms using low-degree polynomials, achieving competitive accuracy with minimal training overhead; and
(2) a generalized interleaved packing (GIP) scheme that is compatible with feature maps of virtually arbitrary spatial resolutions, accompanied by a suite of carefully designed homomorphic operators that preserve the GIP-form encryption throughout computation.
These advances enable efficient, end-to-end FHE inference across diverse CNN architectures.
Experiments on CIFAR-10, ImageNet, and MS COCO demonstrate that the FHE-friendly CNNs obtained via our SFT strategy achieve accuracy comparable to baselines using ReLU or SiLU activations. Moreover, this work presents the first demonstration of FHE-based inference for YOLO architectures in object detection leveraging low-degree polynomial activations.
\end{abstract}

\section{Introduction}

The proliferation of \emph{prediction-as-a-service} (PaaS) introduces significant privacy risks for client data. While \emph{Fully Homomorphic Encryption} (FHE)~\cite{Gentry_2009} offers a solution by enabling computation on encrypted data, deploying Convolutional Neural Networks (CNNs) under FHE remains challenging. The core difficulties lie in approximating non-linear activations like ReLU with polynomials with minimal accuracy degradation, and overcoming the ciphertext capacity barrier that constrains high-resolution image processing on FHE inference.

Existing solutions to non-linear activations primarily employ either high-degree polynomial substitutions~\cite{Lee_2021,Lee_2022,Lee_2023}, which incur high FHE inference latency, or low-degree approximations~\cite{cryptoeprint_2017, Obla_2020, Ishiyama_2020, garimella2021sisyphus, Baruch_2022, Park_2022_AESPA, baruch2023training}, which often suffer from significant accuracy degradation on deep CNNs~\cite{cryptoeprint_2017, Obla_2020, Ishiyama_2020, garimella2021sisyphus, Baruch_2022}. Moreover, low-degree polynomial approximation methods typically necessitate resource-intensive retraining from scratch~\cite{Park_2022_AESPA} or multi-stage fine-tuning~\cite{baruch2023training} with sensitive hyperparameters. Furthermore, conventional inference framework~\cite{Juvekar_2018_Gazelle, Lee_2021, Lee_2022, Kim_2024_HyPHEN} including homomorphic convolutions hit a \emph{size barrier} with high-resolution images, as their per-channel pixel counts exceed the slots available in a CKKS~\cite{Cheon_2017} ciphertext.

To address these challenges, we propose a general framework for adapting various deep CNN architectures to FHE inference. Our contributions are twofold:
(1) a \emph{single-stage fine-tuning (SFT)} strategy that directly converts pre-trained CNNs into FHE-friendly models using low-degree (e.g., degree-4) polynomials, achieving competitive accuracy with minimal training overhead; and
(2) a \emph{generalized interleaved packing (GIP)} scheme that is compatible with feature maps of virtually arbitrary spatial resolutions, accompanied by a suite of carefully designed homomorphic operators (e.g., convolution and deconvolution) that preserve the GIP-form encryption throughout computation.
These advances enables efficient end-to-end FHE inference across diverse CNN architectures, including ResNet, MobileNet, YOLOv5~\cite{Jocher_YOLOv5_2020}, and U-Net~\cite{ronneberger2015u}.

Experiments on CIFAR-10, ImageNet, and MS COCO datasets demonstrate that the FHE-friendly CNNs obtained via our SFT strategy achieve competitive accuracy compared to baselines with activations such as ReLU or SiLU. Moreover, to the best of our knowledge, this work presents the first demonstration of FHE inference for YOLO architectures (e.g., YOLOv5) for object detection leveraging low-degree (degree-4) polynomial activations.

\section{Related Work}

\subsection{CKKS Fully Homomorphic Encryption}
\label{subsec:fhe_background}

CKKS~\cite{Cheon_2017} is a homomorphic encryption scheme that supports fixed-point arithmetic on encrypted data. In this scheme, a vector message $\mathbf{m} \in \mathbb{R}^{N/2}$ (or $\mathbb{C}^{N/2}$) is first encoded into a plaintext polynomial in the cyclotomic ring $\mathcal{R}_Q = \mathbb{Z}_Q[X]/(X^N + 1)$ for a power-of-two $N$, and then encrypted into a ciphertext—a pair of polynomials that conceals the message with noise. A key feature of CKKS is \textit{slot batching}, which enables a single homomorphic operation such as addition or multiplication to be applied component-wise to all $N/2$ elements of the underlying vector in one step. Each position in this vector is referred to as a \textit{slot}.

The core homomorphic operations in CKKS are summarized below, where $[\mathbf{u}]$ denotes the ciphertext encrypting vector $\mathbf{u}$:
\begin{itemize}
\item \textbf{Addition and Multiplication:} Both homomorphic addition and multiplication can be performed between two ciphertexts or between a ciphertext and a plaintext. Addition and multiplication are executed in a component-wise manner.
\item \textbf{Rotation:} This operation cyclically shifts the slots within a ciphertext. A rotation by $r$ positions shifts the vector left when $r>0$ and right when $r<0$.
\end{itemize}

A major limitation in homomorphic encryption is the restricted capacity for sequential multiplications. When a ciphertext's level is exhausted ($\ell=0$), a computationally expensive process called \textit{bootstrapping} is required to refresh the ciphertext. The incorporation of bootstrapping makes CKKS a fully homomorphic encryption scheme~\cite{cheon2018bootstrapping}, enabling an unlimited number of multiplications on encrypted data.

\subsection{Polynomial Activation}
\label{subsec:poly_activation}

A central challenge in constructing FHE-friendly CNNs lies in replacing non-polynomial activation functions, such as ReLU or SiLU, with polynomial approximations. Existing approaches to this problem can be broadly categorized into two classes.

The first strategy employs high-degree polynomials to approximate activation functions directly~\cite{Lee_2021,Lee_2022,Lee_2023}. Although this approach maintains accuracy without requiring model retraining, it introduces substantial computational overhead in FHE inference due to the significant cost of multiplicative depth, resulting in frequent and costly bootstrapping operations. Moreover, such methods are highly sensitive to input ranges, as the approximation degrades significantly when inputs fall outside the predefined interval.

The second strategy involves retraining or fine-tuning CNNs with low-degree polynomial activations. These methods face a critical trilemma involving accuracy, efficiency, and practicality: they often suffer from (i) poor trainability and severe accuracy degradation in deep CNNs, and (ii) incompatibility with one-stage fine-tuning strategies that aim to produce FHE-friendly models with minimal overhead—for instance, those requiring an order of magnitude fewer training epochs. For example, the method in~\cite{Baruch_2022} that gradually replaces ReLU with polynomial activations exhibits notable accuracy loss. Similarly, the method in \cite{garimella2021sisyphus} that aligns intermediate outputs of a polynomial model with those of a pre-trained network using ReLU activations, fails to bridge the accuracy gap. The method in~\cite{Park_2022_AESPA} replaces ReLU and Batch Normalization with a composition of orthogonal basis polynomials and basis-wise batch normalization. Although the resulting representation can be fused into a simple square function during inference—significantly accelerating FHE inference—the fusion of orthogonal bases and basis-wise normalization may distort the distribution of intermediate features, thereby impeding effective fine-tuning from pre-trained models.

To mitigate training instability, the method in~\cite{baruch2023training} introduces an auxiliary loss term that constrains the input magnitudes to polynomial activation layers. This underscores the importance of controlling activation input ranges: a smaller interval allows lower-degree polynomials to achieve comparable accuracy, thus improving FHE inference efficiency. However, the method in \cite{baruch2023training} requires carefully balancing the range constraint loss and the task-specific loss via a weighting parameter, which is difficult to set a priori. An excessive weight may dominate the learning objective and hinder convergence, whereas an insufficient one may fail to adequately bound activation inputs. Moreover, the approach relies on multi-stage fine-tuning with sensitive hyperparameters, complicating its practical deployment and limiting its generalizability across CNN architectures.

These limitations collectively highlight the need for a streamlined and robust approach that can adapt pre-trained CNNs into stable and FHE-efficient models using low-degree polynomials—without multi-stage procedures or significant accuracy degradation. Our work addresses this gap by introducing a \textit{single-stage fine-tuning} strategy.

\subsection{Homomorphic Convolution on Encrypted Data}

Gazelle~\cite{Juvekar_2018_Gazelle} introduces a pioneering convolutional algorithm tailored for Homomorphic Encryption (HE) that avoids costly FHE-based data rearrangement by integrating secure multi-party computation (MPC). Subsequent work~\cite{Lee_2021} extends Gazelle's approach by enabling FHE evaluation of data rearrangement operations, supporting end-to-end FHE inference. Additionally, the method in~\cite{Lee_2021} develops a multiplexed packing scheme that employs denser data formats to minimize ciphertext requirements. HyPHEN~\cite{Kim_2024_HyPHEN} introduces flexible data formats to streamline inter-convolution data arrangement and proposes row-wise image segmentation to reduce plaintext weight size during convolution.

These implementations face fundamental limitations in high-resolution scenarios where the pixel count per channel exceeds the available slot capacity of a single CKKS ciphertext. The multiplexed packing method~\cite{Lee_2021} only functions when per-channel pixel counts remain within slot capacity, while HyPHEN's non-overlapping row segmentation exclusively supports $1\times 1$ convolution filters. For larger filters, segments require overlapping rows to maintain correctness at boundaries. However, the required number of overlapping rows increases progressively through successive convolution layers, necessitating meticulous pre-calculation across the entire network architecture. This requirement becomes particularly burdensome for deep CNNs. Moreover, extensive overlapping leads to severe ciphertext slot under-utilization, degrading computational efficiency.

Gazelle employs interleaved decomposition to handle strided convolutions, distributing input pixels across multiple ciphertexts to decompose strided convolution into sums of stride-1 convolutions. However, this approach relies on multi-party computation (MPC) to rearrange outputs for subsequent convolutional layers. Furthermore, the method does not support stride-1 convolutions on high-resolution images. These limitations constrain its applicability in general scenarios.

Collectively, this spatial resolution barrier imposed by ciphertext slot capacity severely restricts practical deployment of FHE-based CNNs for real-world high-resolution image data. We address this challenge through a Generalized Interleaved Packing (GIP) scheme that supports feature maps of virtually arbitrary spatial resolutions, accompanied by carefully designed homomorphic operators including convolution that maintain the GIP-form encryption throughout computation.

\section{Methodology}

\subsection{Single-Stage Fine-Tuning}

To fine-tune a pretrained CNN model into a FHE-friendly form, non-polynomial activations need to be accurately approximated by their polynomial counterparts. Since the polynomial approximation is only valid within a bounded input range, we introduce the following \emph{homogeneous property} of the ReLU activation function:
\begin{align}
\label{eq:scale rule}
\operatorname{ReLU}(x) = q \cdot \operatorname{ReLU}\left(\frac{x}{q}\right), \quad \forall x \in \mathbb{R},\ q > 0.
\end{align}
Building upon this property, the method in~\cite{Lee_2023} approximates ReLU by a polynomial function $\operatorname{poly}(x)$ as:
\begin{align}
\label{eq:approx rule}
\operatorname{ReLU}(x) \approx q \cdot \operatorname{poly}\left(\frac{x}{q}\right).
\end{align}

However, the method in~\cite{Lee_2023} requires a pre-defined global maximum value $B$ (corresponding to $q$ in Equation (\ref{eq:approx rule})) for all activation inputs. 
Additionally, even with fine-tuning, it relies on high-degree polynomials (degree 637 for $\alpha=10$) to avoid significant accuracy degradation on CIFAR-10. 
These requirements substantially limit its practical applicability across diverse datasets and architectures.

To overcome these limitations, we introduce a \emph{single-stage fine-tuning} strategy (Algorithm~\ref{alg:single-stage-finetuning}) centered around our novel PolyAct-RN operator (Algorithm~\ref{alg:polyact-rn}). Unlike the approach in~\cite{Lee_2023} that does not constrain activation input ranges, PolyAct-RN adaptively normalizes activation inputs channel-wise to a consistent range $[-\gamma, \gamma]$ with $\gamma>0$. Within this constrained range, PolyAct-RN employs a degree-$d$ polynomial to approximate the activation function, then rescales the output to restore the original magnitude, following the homogeneous principle in Equation (\ref{eq:scale rule}). This approach maintains model accuracy while using significantly lower-degree polynomials than the approach in~\cite{Lee_2023}, offering substantial efficiency gains for FHE inference. 

In all experiments, we set $\gamma = 3$ and the polynomial degree $d = 4$. Specifically, we compute the degree-4 polynomial $\operatorname{poly}(x)$ as a weighted sum of orthonormal Hermite polynomials following~\cite{Park_2022_AESPA}: $\operatorname{poly}(x) = \sum_{i=0}^4 \hat{f}_i h_i(x)$, with basis functions and coefficients provided in Table~\ref{tab:approximate_coefficients}. Alternative approximation methods for obtaining $\operatorname{poly}(x)$ could also be employed. 

It is noteworthy that during inference, PolyAct-RN reduces to a fixed fourth-degree polynomial whose coefficients are independent of the input, making it fully compatible with FHE evaluation.

\begin{algorithm}[!t]
\caption{PolyAct-RN: Polynomial Activation with Range Normalization}
\label{alg:polyact-rn}
\begin{algorithmic}[1]
\REQUIRE Mini-batch input tensor $X \in \mathbb{R}^{B \times C \times H \times W}$, approximation range parameter $\gamma$, degree-$d$ polynomial $\operatorname{poly}(\cdot)$, momentum parameter $\beta$, initialized running statistics $\{M_c^{\text{inf}}=1\}_{c=1}^C$, numerical stability constant $\epsilon$
\ENSURE Output tensor $Y \in \mathbb{R}^{B \times C \times H \times W}$

\IF{mode = training}
    \STATE Compute maximum absolute value: $M_c = \max_{b,h,w} |X_{b,c,h,w}|$, $c = 1$, 2, $\ldots$, $C$
    \STATE Update running statistic: $M_c^{\text{inf}} = \beta \cdot M_c^{\text{inf}} + (1 - \beta) \cdot M_c$, $c = 1$, 2, $\ldots$, $C$
\ENDIF

\FOR{$c = 1$ to $C$}
    \STATE $q_c \leftarrow \begin{cases} \frac{M_c}{\gamma} + \epsilon & \text{if training} \\ \frac{M_c^{\text{inf}}}{\gamma} + \epsilon & \text{if inference} \end{cases}$
    
    \FOR{$b = 1$ to $B$, $h = 1$ to $H$, $w = 1$ to $W$}
        \STATE Normalize: $\hat{Y}_{b,c,h,w} = \frac{X_{b,c,h,w}}{q_c}$
        \STATE Evaluate polynomial: $\tilde{Y}_{b,c,h,w} = \operatorname{poly}(\hat{Y}_{b,c,h,w})$
        \STATE Rescale: $Y_{b,c,h,w} = q_c \cdot \tilde{Y}_{b,c,h,w}$
    \ENDFOR
\ENDFOR
 \RETURN $Y$
\end{algorithmic}
\end{algorithm}

\begin{algorithm}[!t]
\caption{Single-Stage Fine-Tuning}
\label{alg:single-stage-finetuning}
\begin{algorithmic}[1]
\REQUIRE Pre-trained CNN model $\mathcal{M}$, dataset $\mathcal{D}$
\ENSURE FHE-friendly CNN model $\mathcal{M}'$
\STATE Replace all activation functions in $\mathcal{M}$ with PolyAct-RN defined in Algorithm \ref{alg:polyact-rn}
\STATE Replace all MaxPool layers in $\mathcal{M}$ with AvgPool layers
\STATE If necessary, resize input images to the target resolution
\STATE Fine-tune the modified model $\mathcal{M}'$ on $\mathcal{D}$ for a small number of epochs
\RETURN $\mathcal{M}'$
\end{algorithmic}
\end{algorithm}

\begin{figure}[!t]
  \centering
  \includegraphics[width=0.7\textwidth]{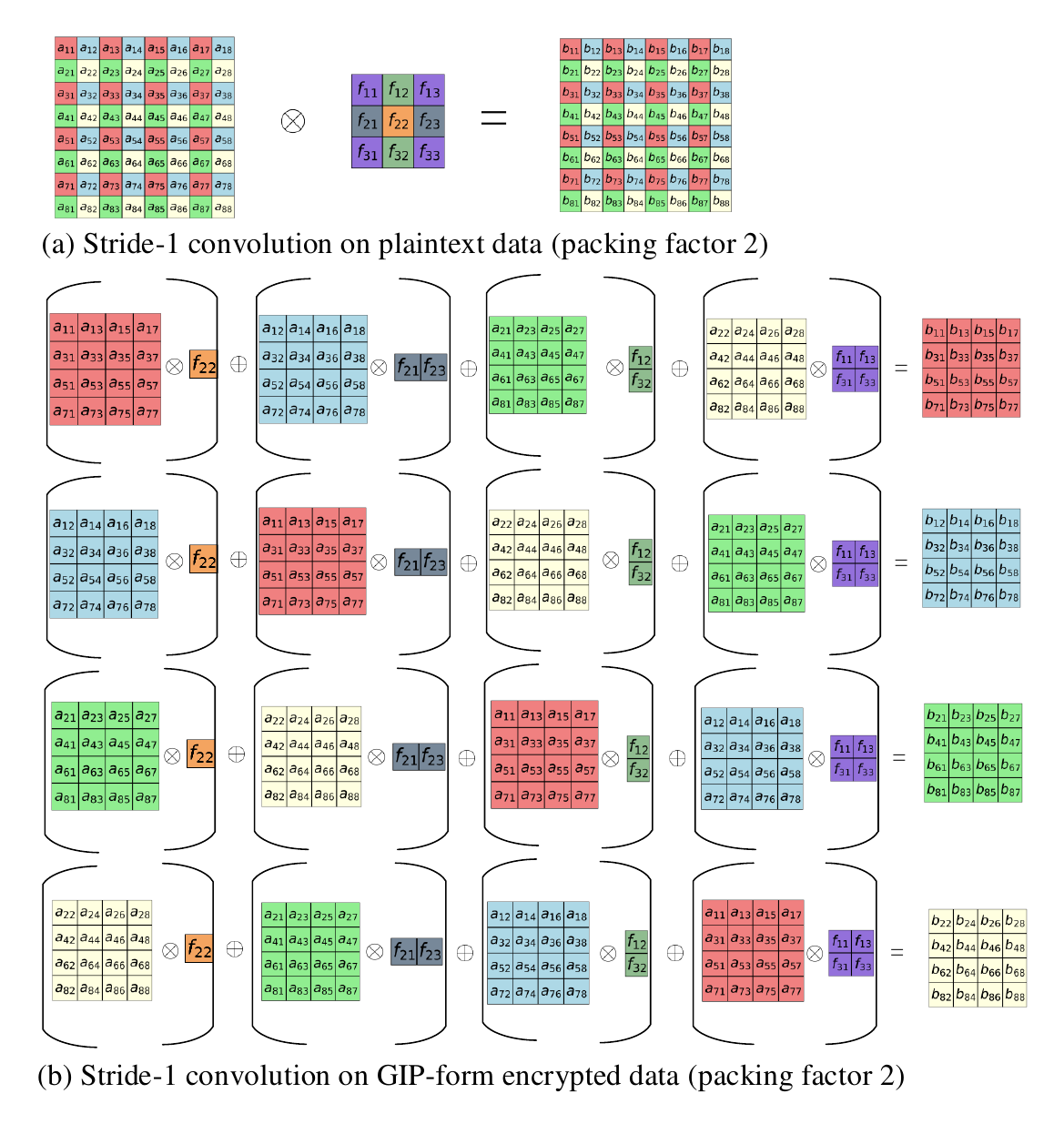}
  \caption{Illustration of the proposed homomorphic convolution. (a) Standard convolution performed on plaintext data with a channel packing factor of 2. (b) Homomorphic convolution executed on the corresponding GIP-form encrypted data. For clarity, we depict the single input–output channel case.}
  \label{fig:fig1}
\end{figure}

\subsection{Generalized Interleaved Packing for FHE Inference}

We introduce the \emph{Generalized Interleaved Packing} (GIP) scheme that is compatible with feature maps of virtually arbitrary spatial resolutions and carefully design homomorphic operators—including convolution, deconvolution, and average pooling—that take encrypted inputs in GIP form and produce encrypted outputs in GIP form. We employ \texttt{torch.nn.Conv2d} and \texttt{torch.nn.ConvTranspose2d} from PyTorch~\cite{NEURIPS2019_pytorch} as the functional definitions of convolution and deconvolution operators, respectively.
Moreover, during FHE inference, the GIP scheme remains compatible with component-wise operators such as batch normalization and polynomial activations. By combining these advances, GIP enables end-to-end FHE inference for diverse CNN architectures, including YOLOv5~\cite{Jocher_YOLOv5_2020} and U-Net~\cite{ronneberger2015u}, with both high computational efficiency and structural flexibility.

We define the \emph{channel packing factor} for a feature map as $g = H/\hat{H}$,
where $H\times H$ denotes the spatial dimensions (height and width) of the feature map (assumed square for simplicity), and $\hat{H}^2$ denotes the base packing size, which does not exceed the slot capacity of a CKKS ciphertext.
In our generalized interleaved packing scheme, given a homomorphic operator at the $i$-th layer of a CNN with input channel packing factor $g_{i}$, the output channel packing factor $g_{i+1}$ is determined as follows:
$g_{i+1}=g_i / s_i$ for down-sampling operators (strided convolution or strided average pooling) with stride $s_i$;
$g_{i+1}=g_i \cdot \hat{s}_i$ for up-sampling operators (deconvolution or nearest-neighbor upsampling) with stride $\hat{s}_i$;
and $g_{i+1}=g_i$ for resolution-preserving operators (batch normalization, polynomial activations, stride-1 convolution, or stride-1 average pooling).
Additionally, GIP adaptively selects the packing scheme based on the channel packing factor $g_i$.
When $g_i > 1$, it employs interleaved decomposition packing~\cite{Juvekar_2018_Gazelle}, where pixels from a single channel are distributed across $g_i^2$ ciphertexts, with interleaved pixels of the same color packed into the same ciphertext as illustrated in Figure~\ref{fig:fig1}.
Conversely, when $g_i < 1$, GIP adopts multiplexed packing~\cite{Lee_2021}, where pixels from $1/g_i^2$ channels are interleaved and packed into a single ciphertext, and the gap between pixels of the same channel within a ciphertext is $1/g_i$. 
When $g_i = 1$, the two schemes coincide, allowing seamless transitions between them without any data rearrangement.
This design enables efficient application of down-sampling, up-sampling, and size-preserving operators across feature maps of different resolutions.

Finally, we describe the design of our homomorphic convolution with stride $s_i\geq 1$, which takes inputs with a channel packing factor $g_i > 1$ and produces outputs with a channel packing factor $g_{i+1} \geq 1$. 
Figure~\ref{fig:fig1} illustrates an example with stride 1, padding 1, and $3 \times 3$ kernels, where both input and output channels have a packing factor of 2. 
In such a configuration, existing homomorphic convolution methods encounter a \emph{size barrier}, as the per-channel pixel count of both input and output exceeds the available slots in a CKKS ciphertext. 
In Figure~\ref{fig:fig1} (b), the input and output channels are each decomposed into four sub-channels in an interleaved manner, with each sub-channel packed into a separate CKKS ciphertext. Each output sub-channel is computed by aggregating stride-1 convolutions between the corresponding input sub-channels and their respective, interleaved kernel coefficients. Each of these core stride-1 convolutions is implemented in the \emph{same} style, utilizing zero-padding on the input to ensure the output resolution matches the input.   
Other convolution configurations could be implemented accordingly. Notably, Gazelle~\cite{Juvekar_2018_Gazelle} decomposes strided convolutions into sums of stride-1 convolutions but requires $g_i = s_i$ and $g_{i+1} = 1$, necessitating multi-party computation (MPC) to rearrange outputs for subsequent convolutions. In contrast, our homomorphic convolution only requires $g_{i+1}=g_i / s_i$, enabling end-to-end FHE inference without MPC.

\begin{table}[!t]
 \caption{Comparison of CNNs and their FHE-friendly counterparts on the CIFAR-10 or ImageNet datasets}
 \label{tab:CIFAR_ImageNet_comparison}
  \centering
  \begin{tabular}{lccccc}
    \toprule
     Model & Dataset & Input Size & Params (M) & Accuracy (Baseline) & Accuracy (FHE-friendly)\\
     \midrule 
     ResNet20 & CIFAR-10 & $32^2$ & 0.3 & 0.919 & 0.920\\
     ResNet32 & CIFAR-10 & $32^2$ & 0.5 & 0.929 & 0.931\\
     ResNet44 & CIFAR-10 & $32^2$ & 0.6 & 0.935 & 0.936 \\
     ResNet18 & ImageNet & $224^2$ & 11.7 & 0.698 & 0.678\\
     MobileNetv1 & ImageNet & $224^2$ & 4.2 & 0.707 & 0.687 \\
     MobileNetv2 & ImageNet & $224^2$ & 3.5 & 0.719 & 0.701\\
    \bottomrule
  \end{tabular}
\end{table}

\begin{table}[!t]
 \caption{Comparison of YOLOv5 variants and their FHE-friendly counterparts on MS COCO}
 \label{tab:yolov5_comparison}
  \centering
  \begin{tabular}{lccccccc}
    \toprule
    Model & Dataset & Input Size & Params (M) & \multicolumn{2}{c}{mAP@0.5:0.95} & \multicolumn{2}{c}{mAP@0.5}\\
    \cmidrule(lr){5-6} \cmidrule(lr){7-8} 
    & & & & Baseline & FHE-friendly & Baseline & FHE-friendly \\
    \midrule
    YOLOv5n & MS COCO & $640^2$ & 1.9 & 0.280 & 0.257 & 0.457 & 0.427 \\
    YOLOv5s & MS COCO & $640^2$ & 7.2 & 0.374 & 0.352 & 0.568 & 0.542 \\
    YOLOv5m & MS COCO & $640^2$ & 21.2 & 0.454 & 0.433 & 0.641 & 0.614\\
    YOLOv5l & MS COCO & $640^2$ & 46.5 & 0.490 & 0.463 & 0.673 & 0.642\\
    YOLOv5x & MS COCO & $640^2$ & 86.7 & 0.507 & 0.477 & 0.689 & 0.650\\
    \bottomrule
  \end{tabular}
\end{table}

\begin{table}[!t]
 \caption{FHE-based inference performance of YOLOv5n and MobileNetv2 using a single CPU thread}
 \label{tab:yolov5 FHE inference}
  \centering
  \begin{tabular}{lccccc}
    \toprule
    Model & Dataset & Input Size & Params (M) &  Metric & FHE Inference Time ($s$)\\
    \midrule
    YOLOv5n & MS COCO & $512^2$ & 1.9 &  0.241 (mAP@0.5:0.95) & 8889.9 \\
    MobileNetv2 & ImageNet & $256^2$ & 3.5 & 0.705 (Accuracy) & 11258.9  \\
    \bottomrule
  \end{tabular}
\end{table}

\section{Evaluation}

We evaluate our single-stage fine-tuning (SFT) strategy, as detailed in Algorithm~\ref{alg:single-stage-finetuning}, on three benchmark datasets: CIFAR-10~\cite{krizhevsky2009learning}, ImageNet~\cite{Deng_2009_ImageNet}, and MS COCO~\cite{lin2014microsoft}. Our experimental setup encompassed ResNet-20/32/44 architectures on CIFAR-10, ResNet-18, MobileNetv1, and MobileNetv2 on ImageNet, and various YOLOv5~\cite{Jocher_YOLOv5_2020} variants on MS COCO.

The CIFAR-10 dataset comprises 60,000 color images distributed across 10 classes, with a standard split of 50,000 training and 10,000 validation images. ImageNet contains approximately 1.28 million training images and 50,000 validation images across 1,000 classes. For object detection tasks, we employed the MS COCO dataset, which includes 118,287 training images and 5,000 validation images over 80 categories.

We compare baseline models using standard activations (e.g., ReLU/SiLU) and MaxPool layers with their FHE-friendly counterparts obtained via our SFT strategy (Algorithm~\ref{alg:single-stage-finetuning}). Results are summarized in Tables~\ref{tab:CIFAR_ImageNet_comparison} and~\ref{tab:yolov5_comparison}.
As shown, FHE-friendly CNNs achieve accuracies comparable to their baseline counterparts on CIFAR-10 and ImageNet datasets, while FHE-friendly YOLOv5 variants maintain competitive object detection performance in terms of mAP@0.5:0.95 and mAP@0.5.

Furthermore, Table~\ref{tab:yolov5 FHE inference} presents the FHE inference performance of the YOLOv5n and MobileNetv2 models. 
Their FHE-friendly versions are derived using the proposed SFT strategy, with input images resized to powers of two to align with CKKS ciphertext slot capacities. The baseline architecture of YOLOv5 (excluding the detection head) is detailed in the official documentation\footnote{\url{https://docs.ultralytics.com/yolov5/tutorials/architecture_description/\#1-model-structure}}.
All experiments in~Table~\ref{tab:yolov5 FHE inference} were performed on an Ubuntu 20.04.1 LTS (64-bit) system equipped with an
Intel\textregistered\ Xeon\textregistered\ Gold 6226R CPU@2.90 GHz with 375 GB memory, using a single CPU thread. The CKKS encryption parameters satisfy a 128-bit security level with a polynomial degree of $N = 2^{16}$, resulting in a ciphertext slot capacity of  $2^{15}$.

As summarized in Table~\ref{tab:yolov5 FHE inference}, the end-to-end FHE inference on YOLOv5n (excluding the detection head) requires 8889.9 seconds using a single CPU thread. 
This runtime could be significantly reduced through hardware acceleration using GPUs or FPGAs, potentially by around three orders of magnitude.

\begin{table}[!t]
 \caption{Hermite polynomials and weighted coefficients for degree-4 polynomial approximations of ReLU and SiLU}
  \label{tab:approximate_coefficients}
  \centering
  \begin{tabular}{lccccc}
    \toprule
      & $i=0$ & $i=1$ & $i=2$ & $i=3$ & $i=4$\\
     \midrule 
      Basis function $h_i(x)$ & $1$ & $x$ & $\frac{x^2-1}{\sqrt{2}}$ & $\frac{x^3-3x}{\sqrt{6}}$ & $\frac{x^4 - 6x^2 + 3}{2\sqrt{6}}$ \\
      \midrule
     Coefficient $\hat{f}_i$ (ReLU) & 0.39894228 & 0.5 & 0.28209479 & 0 & -0.08143375\\
     \midrule
     Coefficient $\hat{f}_i$ (SiLU) & 0.20662096 & 0.5 & 0.24808519 & 0 & -0.03780501\\
    \bottomrule
  \end{tabular}
\end{table}

\section{Conclusion}

We present a single-stage fine-tuning (SFT) strategy with low-degree polynomial activations to convert diverse CNNs into FHE-friendly forms, achieving competitive accuracy with minimal training overhead. Combined with our Generalized Interleaved Packing (GIP) scheme and carefully designed homomorphic operators, this framework enables efficient end-to-end FHE inference across these architectures. To the best of our knowledge, this work presents the first demonstration of FHE inference for YOLO architectures in object detection using low-degree (degree-4) polynomial activations.

\printbibliography

\end{document}